\documentclass[10pt, onecolumn]{IEEEtran}

\usepackage[english]{babel}
\usepackage{enumitem}
\usepackage[latin1]{inputenc}
\usepackage{enumerate}
\usepackage{graphicx}
\usepackage{color}
\usepackage[dvipsnames]{xcolor}
\usepackage[T1]{fontenc}
\usepackage{dsfont}
\usepackage{amsfonts}
\usepackage{graphicx,wrapfig}
\usepackage[T1]{fontenc}
\usepackage{amsmath}
\usepackage{mathtools, cuted}
\usepackage{amsthm}
\usepackage{amstext}
\usepackage{amssymb}
\usepackage{mathrsfs}
\usepackage{mathtools}
\usepackage{tikz}
\usepackage{marginnote}
\usepackage{tkz-tab}
\usetikzlibrary{topaths,calc}
\usepackage{stackengine}

\def\R{\mathbb{R}}

\def\argmax{\mathop{\rm arg\, max}}
\def\argmin{\mathop{\rm arg\, min}}

\def\D{{\mathcal D}}

\def\P{{\mathcal P}}

\usepackage[font=small,labelsep=space]{caption}
\DeclareCaptionLabelSeparator{dot}{.~}
\newcounter{example}

\newtheorem{definition}{Definition}
\newtheorem{theorem}{Theorem}

\newtheorem{proposition}{Proposition}
\newtheorem{lemma}{Lemma}
\theoremstyle{remark}
\newtheorem{remark}{Remark}

\usepackage{times}
\usepackage{tikz}
\usepackage{amsmath}
\usepackage{verbatim}
\usetikzlibrary{arrows,shapes}
\tikzstyle{RectObject}=[rectangle,fill=white,draw,line width=0.2mm]
\tikzstyle{line}=[draw]
\tikzstyle{arrow}=[draw, -latex]
\usetikzlibrary{decorations.pathmorphing}
\usetikzlibrary{calc,shapes, positioning}
\usepackage{graphicx}
\usepackage{caption}
\usetikzlibrary{shapes.geometric}

\definecolor{DukeBlue}{HTML}{001A57}
\definecolor{DarkRed}{rgb}{0.75, 0.0, 0.0}
\definecolor{DarkGreen}{rgb}{0.0, 0.5, 0.0}

\DeclareFontFamily{U}{BOONDOX-calo}{\skewchar\font=45 }
\DeclareFontShape{U}{BOONDOX-calo}{m}{n}{
	<-> s*[1.05] BOONDOX-r-calo}{}
\DeclareFontShape{U}{BOONDOX-calo}{b}{n}{
	<-> s*[1.05] BOONDOX-b-calo}{}
\DeclareMathAlphabet{\mathcalboondox}{U}{BOONDOX-calo}{m}{n}
\SetMathAlphabet{\mathcalboondox}{bold}{U}{BOONDOX-calo}{b}{n}
\DeclareMathAlphabet{\mathbcalboondox}{U}{BOONDOX-calo}{b}{n}

\allowdisplaybreaks

\newcommand{\paren}[1]{\left(#1\right)}

\usepackage[ruled]{algorithm2e} 
\usepackage{hyperref, xr, subfig}

\date{}

\title{\LARGE \bf
	Wasserstein Soft Label Propagation on Hypergraphs: Algorithm and Generalization Error Bounds
}

\author{Tingran Gao$^{1}$, Shahab Asoodeh$^{2}$, Yi Huang$^{3}$, and James Evans$^{4}$
    	\thanks{$^{1}$Department of Statistics,
		The University of Chicago, Chicago, IL 60637
		{\tt\small tingrangao@galton.uchicago.edu}}%
    	\thanks{$^{2}$Computation Institute and Institute of Genomics and System Biology, The University of Chicago, Chicago, IL 60637 
		{\tt \small A@uchicago.edu}}
       \thanks{$^{3}$Department of Medicine, University of Chicago, Chicago IL, {\tt\small yhuang10@uchicago.edu}}
		\thanks{$^{4}$Computation Institute and Department of Sociology, The University of Chicago, Chicago, IL 60637
		{\tt\small jevans@uchicago.edu}}
}

\begin{document}
	\maketitle

	\begin{abstract}
		Inspired by recent interests in developing machine learning and data mining algorithms for hypergraphs, here we investigate the semi-supervised learning algorithm of propagating "soft labels" (e.g. probability distributions, class membership scores) over hypergraphs, by means of optimal transportation. Borrowing insights from Wasserstein propagation on graphs [Solomon et al. 2014], we re-formulate the label propagation procedure as a message-passing algorithm, which renders itself naturally to a generalization applicable to hypergraphs through Wasserstein barycenters. 
		Furthermore, in a PAC learning framework, we provide generalization error bounds for propagating one-dimensional distributions on graphs and hypergraphs using 2-Wasserstein distance, by establishing the \textit{algorithmic stability} of the proposed semi-supervised learning algorithm. These theoretical results also offer novel insight and deeper understanding about Wasserstein propagation on graphs. 
	\end{abstract}
	
	\section{Introduction}
	Recent decades have witnessed a growing interest in developing machine learning and data mining algorithms on \emph{hypergraphs} \cite{Hypergraph_Clustering,Hypergraph_Jost,Hypergraph_Game,Hypergraph_Kannan,Hypergraph_Olgica,Hypergraph_TV,HypergraphScalable}. As a natural generalization of graphs, a hypergraph is a combinatorial structure consisting of vertices and hyperedges, where each hyperedge is allowed to connect any number of vertices. This additional flexibility facilitates the capture of higher order interactions among objects; applications have been found in many fields such as computer vision \cite{Govindu2005}, network clustering \cite{Hyper_Spatial}, folksonomies \cite{GZCN2009}, cellular networks \cite{KHT2009}, and community detection \cite{KBG2018}.
	
	This paper develops a probably approximately correct (PAC) learning framework for \emph{soft label propagation} or \emph{Wasserstein propagation} \cite{Solomon:2014}, a recently proposed semi-supervised learning algorithm based on optimal transport \cite{villani2003topics,villani2008optimal}, on graphs and hypergraphs. Distinct from the prototypical semi-supervised learning algorithm of \emph{label propagation} \cite{Belkin2004}, in which labels of interest are numerical or categorical variables, Wasserstein propagation aims at inferring unknown \emph{soft labels}, such as histograms or probability distributions, from known ones, based on pairwise similarities qualitatively characterized by edge connectivity and quantitatively measured using Wasserstein distances. Compared with traditional ``hard labels,'' soft labels are built with extra flexibility and informativeness, rendering themselves naturally to applications where uncertainty and distributional information is crucial. For example, the traffic density at routers on the Internet network or topic distributions across the co-authorship network are more naturally modeled as probability distributions.
	
	Semi-supervised learning is a paradigm that leverages unlabelled data to improve the generalization performance for supervised learning, under generic, unsupervised structural assumptions about the dataset (e.g. the manifold assumption); see \cite{Seeger01learningwith,Zhu06SSL,CSZ2006} for an overview. Given a graph $G=\left( V,E \right)$ and a subset of vertices $V_0\subset V$, label propagation is a procedure for extending an assignment of labels on $V_0$, denoted as a map $f_0:V_0\rightarrow \D$ valued in an arbitrary set $\D$, to a map $f:V\rightarrow\D$ on the entire vertex set $V$. Borrowing an analogy from the classical heat equation, this extension procedure is reminiscent of heat propagation from ``boundary'' $V_0$ to the ``entire domain'' $V$. For soft label propagation, the label set $\D$ is the probability distribution $\P \left( N \right)$ modeled on a complete, separable metric space $\left( N,d_N \right)$. 
	
	Among the first works to address semi-supervised learning with soft labels are \cite{Distribution_Propagation1,Distribution_Propagation2,SSL_Measure_Propagation}. In all of these works, the similarity between two soft labels is quantitatively measured using the Kullback-Leibler (KL) divergence, but the soft labels inferred from this process are often unstable and discontinuous. In \cite{Solomon:2014} the authors proposed to replace KL divergence with $1$- or $2$-Wasserstein distance. The resulting soft label propagation algorithm is thus termed ``Wasserstein propagation.'' Specifically, given a measure-valued map $f_0:V_0\rightarrow\P \left( N \right)$ defined on $V_0\subset V$, Wasserstein propagation extends $f_0$ to $f:V\rightarrow \P \left( N \right)$ by solving the variational problem
	\begin{equation}
	\label{eq:wasserstein-propagation}
	\min_{f:V\rightarrow\P \left( N \right)}\sum_{\left( v,w \right)\in E}W_p^p \left( f \left( v \right), f \left( w \right) \right)
	\end{equation}
	subject to the constraint $f\restriction V_0=f_0$. Here $W_p \left( \mu,\nu \right)$ denotes the $p$-Wasserstein distance between probability distributions $\mu,\nu\in\P \left( N \right)$ defined as
	\begin{equation}
	\label{eq:wasserstein-distance}
	W_p \left( \mu,\nu \right)\coloneqq\inf_{\pi\in\Pi \left( \mu,\nu \right)}\left[\iint_{N\times N}d^p_N \left( x,y \right)\,\mathrm{d}\pi \left( x,y \right)\right]^{\frac{1}{p}}
	\end{equation}
	where $\Pi \left( \mu,\nu \right)$ is the set of all probabilistic couplings on $N\times N$ with $\mu$ and $\nu$ as marginals. When $p=2$, the minimizer of \eqref{eq:wasserstein-propagation} can be interpreted as a harmonic map, with boundary condition $f\restriction V_0=f_0$ that takes value in a weak, metric-measure space sense \cite{Otto2001,ambrosio2005gradient,LottVillani2009,Dirichlet_Wasserstein}. Note that this is a nontrivial fact because harmonic maps (or minimizers of the Dirichlet energy) generally only exist when the target metric space $\D$ has negative Alexandrov curvature \cite{Jost}, but $\P \left( N \right)$ equipped with the $2$-Wasserstein distance has positive Alexandrov curvature \cite[\textsection 7.3]{ambrosio2005gradient}. When $\D$ is a one-dimensional distribution on the real line defined by $2$-Wasserstein distance, \cite{Solomon:2014} related \eqref{eq:wasserstein-propagation} to a Dirichlet problem.
	
	In this work, we first extend the framework of \cite{Solomon:2014} to hypergraphs using the \textit{Wasserstein barycenter} \cite{Wasserstein_Barycenter,Hypergraph_Asoodeh}. For $2$-Wasserstein distances this is equivalent to solving a \emph{multi-marginal optimal transport} \cite{CE2010} problem with a naturally constructed cost function. The hypergraph extension of Wasserstein propagation is based on a novel interpretation of the original algorithm on graphs \cite{Solomon:2014} as a message-passing algorithm. Next, we take a deeper look at the statistical learning aspects of our proposed algorithm, and establish generalization error bounds for propagating one-dimensional distributions on graphs and hypergraphs using the $2$-Wasserstein distance. One dimensional distributions such as histograms are among the most frequent applications for soft label propagation. The main technical ingredient is \textit{algorithmic stability} \cite{Algorithmic_Stability}. To our knowledge, our generalization bound is the first of its type in the literature on Wasserstein distance-based soft label propagation; on graphs these results generalize the error bounds from \cite{Belkin2004}. As no general semi-supervised learning algorithm is available for large datasets \cite{Label_Propa_100}, this new connection between the Wasserstein barycenter and semi-supervised learning might be of theoretical as well as computational interest. 
	
	
	In the last section, we provide promising numerical results for both synthetic and real data. In particular, we apply our hypergraph soft label propagation algorithm to random uniform hypergraphs as well as UCI datasets including one on Congressional voting records and another on mushroom characteristics, which are naturally represented using a hypergraph representations.
	
	\subsection{Notation}
	We denote an undirected simple graph as $G=(V, E)$ where $V=[n]\coloneqq\{1, \dots, n\}$ is the vertex set and $E\in V\times V$ denote edges. We use $L$ to denote the (weighted) graph Laplacian associated with (weighted) graph $G$, which is a real square matrix of size $n$-by-$n$ defined by $L:=D-W$, where $W\in \mathbb{R}^{n\times n}$ is the (weighted) adjacency matrix of $G$, and $D\in \mathbb{R}^{n\times n}$ is a diagonal matrix with the (weighted) degree of vertex $j$ at its $\left( j,j \right)$-th entry. We use $H=(V, \mathcal E)$ to denote a hypergraph where $\mathcal E\in 2^V$ is the set of hyperedges of $H$. Given $k\geq 2$ probability measures $\rho_1, \dots, \rho_k$ in $\P(N)$, their \textit{Wasserstein barycenter} is 
	\begin{equation}\label{Def:Barycenter}
	\mathsf{bar}\left(\{\rho_i\}_{i=1}^k\right)\coloneqq \inf_{\nu\in \P(N)}\frac{1}{k}\sum_{i=1}^kW_2^2(\rho_i, \nu).
	\end{equation} 
	Fundamental properties of the minimizer in \eqref{Def:Barycenter} are studied in \cite{Wasserstein_Barycenter}; similar results hold when the squared $2$-Wasserstein distance are weighted differently. Given a hyperedge $E$ of $H$, we use $\mathsf{bar}(E)$ to denote $\mathsf{bar}\left(\{\mu_i\}_{i=1}^{|E|}\right)$ where the probability measures $\mu_1, \dots, \mu_{|E|}$ associated with each vertex $i$ in 
	$E$ are clear from the context.

	\section{Message Passing and Label Propagation on Graphs and Hypergraphs}
	In this section, we formulate our hypergraph label propagation as a special case of belief propagation. To this end, we begin with a brief description of a generalized version of Wasserstein label propagation \cite{Solomon:2014} from a message passing perspective.
	
	A learning problem is specified by a probability distribution $D$ on $X\times Y$  according to which labeled sample pairs $z_i=\left( x_i,y_i \right)$ are drawn and presented to a learning algorithm. The algorithm then outputs a map from $X$ to $Y$. In soft label propagation problems, the maps of interest take values in a space of probability distributions $Y$. From now on, we assume $Y$ is the space of probability distributions on a complete metric space $\left( N,d_N \right)$, i.e., $Y=\P\left( N \right)$. Because $N$ is complete, the space $Y$ equipped with Wasserstein distance is also a complete metric space \cite[Theorem 6.18]{villani2003topics}.
	
	\subsection{Wasserstein Label Propagation on Graphs}
	Let $X$ be a graph $G= \left( V,E \right)$, possibly with weights $\omega_{ij}\geq 0$ on each edge $\left(i,j\right)$. Wasserstein label propagation is an extension of Tikhonov regularization framework on graphs \cite{Belkin2004} from real-valued functions to measure-valued maps. 
	Denote a measure-valued map from $G$ to $\P\left( N \right)$ as $\mu:V\rightarrow\P\left( N \right)$. For simplicity, write $\mu_i:=\mu \left( i \right)$ for $i\in V$. A prototypical semi-supervised learning setting assumes $\mu_1,\cdots,\mu_m$ are known, where $1\leq m \ll n$, and the goal is to determine $\mu_{m+1},\cdots,\mu_n$ on the remaining vertices. We do so by minimizing the following objective function with Tikhonov regularization
	\begin{equation}
	\label{eq:tikhonov}
	\min_{f:V\rightarrow \P\left( N \right)}\frac{1}{m}\sum_{i=1}^m W^2_2 \left( \mu_i,f_i \right)+\gamma\sum_{\left( i,j \right)\in E}\omega_{ij}W^2_2 \left( f_i,f_j \right),
	\end{equation}
	where $\gamma>0$ is a regularization parameter. This minimization problem can be conceived of as an extension of the Dirichlet boundary problem studied in \cite{Solomon:2014} as here we do not impose $f_i=\mu_i$ for $i\in [m]$.
	The minimizer of \eqref{eq:tikhonov} is the measure-valued map ``learned'' from the training data $\left\{ \left( i,\mu_i \right)\mid 1\leq i\leq n \right\}$ and the given graph structure $G= \left( V,E \right)$. We point out that the formulation in \cite{Solomon:2014} is a special case (parameter-free ``interpolated regularization'') of \eqref{eq:tikhonov} in the limit $\gamma\rightarrow0$, for the same reason given in \cite[\textsection 2.2]{Belkin2004}.
	We now provide an algorithm for solving \eqref{eq:tikhonov} based on belief propagation. Because this is only a motivating perspective, we assume for simplicity that the graph is unweighted, but all arguments below can be extended to weighted graphs with heavier notation. In this context, each vertex $i$ updates its \textit{belief} about the local minimizer of \eqref{eq:tikhonov} $f_i$ by exchanging messages to edges to which it is incident. The classical min-sum algorithm \cite{min_Sum} describes this process as follows. 
	At time $t$, vertex $i\in [m]$ has belief $b^{(t)}_i$ about the minimizer $f_i$ of \eqref{eq:tikhonov}; then, at time $t+1$, $i$ sends message $J^{(t)}_{i\to e}$ to edge $e=(i, j)$ and receives message $J^{(t)}_{e\to i}$ from $e$, then updates the message for the next iteration according to
	\begin{equation}\label{Eq:Update2}
	J^{(t)}_{i\to e} \left(b^{(t)}_i\right) = W_2^2\left(\mu_i, b^{(t)}_i\right) + \sum_{k\in N(i)\backslash \{j\}}J^{(t-1)}_{(i,k)\to i}\left(b^{(t-1)}_i\right)
	\end{equation}
	and 
	\begin{equation}\label{Eq:Update}
	J^{(t)}_{e\to i}\left(b^{(t)}_i\right) =  \min_{f_j\in \P(N)}\left[W_2^2\left(b^{(t)}_i, f_j\right) + J^{(t-1)}_{j\to e}\left(f_j\right)\right].
	\end{equation}
	The first term in \eqref{Eq:Update2} is set to zero if $i\notin [m]$. The belief is then updated at time $t+1$ according to evolution
	\begin{equation*}
	b^{(t+1)}_i \coloneqq \argmin_{f_i} \left[W_2(\mu_i, f_i) + \sum_{k\in V: (i, k)\in E } J^{(t)}_{(i, k)\to i}(f_i)\right].
	\end{equation*}
	Convergence of $b_i^{(t)}$ to the true minimizer $f^*_i$ can be guaranteed under mild conditions on initial beliefs if $G$ is a tree (see e.g., \cite{min_Sum}).
	
	
	\subsection{Wasserstein Label Propagation on Hypergraphs}
	Now let $X$ be represented by a hypergraph $H=(V, \mathcal E)$. Because each hyperedge may contain an arbitrary number of vertices, the minimization \eqref{eq:tikhonov} fails to formulate our learning objective. Nevertheless, belief propagation updates \eqref{Eq:Update2} and \eqref{Eq:Update} can naturally be extended passing the message between vertex $i$ and hyperedge $E$ containing $i$ as 
	\begin{equation}\label{Eq:Update_H_2}
	J^{(t)}_{i\to E} \left(b_i^{(t)}\right) =W_2^2\left(\mu_i, b_i^{(t)}\right) +\!\!\!\!\!\!\sum_{E'\in \mathcal E\backslash\{E\}:i\in E'}J^{(t-1)}_{E'\to i}\left(b_i^{(t-1)}\right)
	\end{equation}
	and 
	\begin{equation}\label{Eq:Update_H}
	J^{(t)}_{E\to i}\left(b_i^{(t)}\right) =  \min_{f_{E\backslash \{i\}}}\Big[\mathsf{bar}(E) + \sum_{k\in E\backslash \{i\}}J^{(t-1)}_{k\to E}(f_k)\Big].
	\end{equation}
	where  $f_{E\backslash \{i\}}=\{f_k\in\P(N): k\in E\backslash\{i\}\}$.
	The belief of vertex $i\in [m]$ is then obtained according to the following rule:
	$$b_i^{(t+1)}=\argmin_{f_i\in \P(N)}\left[W_2^2(\mu_i, f_i) + \sum_{E\in \mathcal{E}:i\in E}J^{(t)}_{E\to i}(f_i)\right].$$
	These belief propagation update rules justify the following formulation of label propagation for hypergraphs:
	\begin{equation}\label{eq:tikhonoc_Hyper}
	\min_{f:V\to \P(N)} \frac{1}{m}\sum_{i=1}^m W_2^2\left(\mu_i, f_i\right) + \gamma\sum_{E\in \mathcal E}\mathsf{bar}(E)
	\end{equation}
	which is a natural generalization of \eqref{eq:tikhonov} when the graph is unweighted. For weighted graphs, \eqref{eq:tikhonoc_Hyper} still holds with properly adjusted $\mathsf{bar}(E)$ with weights.
	
	\section{Barycenter and Clique Representation}
	In this section, we assume that labels are one-dimensional probability distributions, i.e., $N\subset \R$, and work solely with the $2$-Wasserstein distance. We will see that in this case, hypergraph label propagation can be cast into a Wasserstein propagation on a weighted graph arising from the clique representation of the hypergraph. The remainder of this paper focuses on establishing generalization error bounds for graphs. The main advantage of one-dimensional soft labels is illustrated by the following classical result in optimal transportation theory.
	\begin{theorem}[\cite{villani2003topics}]\label{Thm:Wasserstein_One}
		Let $\mu, \nu\in \P(N)$ with $N\subset\R$ with cumulative density functions (c.d.f.) $F_\mu$ and $F_\nu$, respectively. Then $$W_2^2(\mu, \nu)=\int_{0}^1\left(F_\mu^{-1}(s)-F_\nu^{-1}(s)\right)^2\text{d}s,$$
		where $F_{\mu}^{-1}$ and $F_{\nu}^{-1}$ are the generalized inverses of $F_\mu$ and $F_\nu$, respectively, i.e., $F_\mu^{-1}(s)\coloneqq \inf\{x\in N:F_\mu(x)>s\}$. 
	\end{theorem}
	The explicit expression for Wasserstein distance enables us to derive the barycenter of any number of one-dimensional distributions in a closed form. 
	\begin{theorem}[\cite{bigot2017}]
		\label{Thm:Barycenter_One}
		Let $\rho_1, \dots, \rho_k\in \P(N)$ be $m$ probability distributions on $N\subset\R$ with cumulative density functions $F_{\rho_i}$, $i\in [k]$. Let $\rho_{\mathsf{b}}$ be the (unique) Wasserstein barycenter of $\{\rho_i\}_{i=1}^k$. Then the generalized inverse c.d.f. $F^{-1}_{\mathsf{b}}$ of $\rho_{\mathsf{b}}$ is given by 
		$$F^{-1}_{\mathsf{b}}(s)=\frac{1}{k}\sum_{i=1}^kF_{\rho_i}^{-1}(s).$$
	\end{theorem}
	Because the inverse cdfs and distributions are in one-to-one correspondence, this theorem characterizes the $2$-Wasserstein barycenter of $\{\rho_i\}_{i=1}^m$. In light of Theorem~\ref{Thm:Barycenter_One}, one can simplify the barycenter of hyperedge $E$ that contains vertices, such as $\{1, 2, \dots, k\}$ as
	\begin{eqnarray}
	\mathsf{bar}(E) &=& \frac{1}{k}\sum_{i=1}^kW^2_2(\mu_i, \mu_\mathsf{b})\nonumber\\
	&=&\frac{1}{k}\sum_{i=1}^k\int_{0}^{1}\left(F_{\mu_i}^{-1}(s)-\frac{1}{k}\sum_{i=1}^kF_{\mu_i}^{-1}(s)\right)^2\text{d}s\nonumber\\
	&=&\frac{1}{k^2}\sum_{i=1}^k\sum_{j=i+1}^n\int_0^1\left(F_{\mu_i}^{-1}(s)-F_{\mu_j}^{-1}(s)\right)^2 \text{d}s \nonumber\\
	&=&\frac{1}{k^2}\sum_{i=1}^k\sum_{j=i+1}^kW_2^2(\mu_i, \mu_j)\label{Eq:One_D_Equivalence}
	\end{eqnarray}
	where the first and second equalities follow from Theorems~\ref{Thm:Wasserstein_One} and \ref{Thm:Barycenter_One}, respectively. Comparing \eqref{Eq:One_D_Equivalence} with \eqref{eq:tikhonoc_Hyper}, we now have
	\begin{proposition}
		\label{prop:clique-equivalence}
		Soft label propagation with $2$-Wasserstein distance for one-dimensional distributions on hypergraphs $H$ using \eqref{eq:tikhonoc_Hyper} is equivalent to Wasserstein propagation on a weighted graph arising from the clique representation $G_H$ of $H$. The weight of each edge $e$ in $G_H$ depends only on the degrees of the hyperedges containing $e$.
	\end{proposition}
	\begin{proof}
		Recall that the \textit{clique representation} of a hypergraph $H=(V, \mathcal E)$ is a graph $G_H=(V, E_H)$, where $E_H=\{(i, j):\exists E\in \mathcal E, \{i, j\}\subset E\}$. The rest of the proof follows from checking definitions.
	\end{proof}
	
	
	\section{Generalization Bounds for Wasserstein Propagation}
	In this section we derive generalization bounds for label propagation \eqref{eq:tikhonov} on graphs. The same results apply to hypergraphs, by Proposition~\ref{prop:clique-equivalence}. We begin by briefly reviewing empirical risk, generalization error, and algorithmic stability in message passing.

	\subsection{Algorithmic Stability}  
	The framework of algorithmic stability \cite{DW1979,Algorithmic_Stability,MNPR2006} was proposed in statistical learning as an alternative to the VC-dimension framework. The latter is often overly pessimistic because it attempts to bound the generalization performance uniformly over all possible algorithms. We briefly recapture the essence of algorithmic stability here. Let $X$ and $Y$ be two measurable spaces, and a set of training samples $S=\left\{ z_i=\left( x_i,y_i \right),i=1,\cdots,m \right\}$ of size $m$ sampled i.i.d. with respect to an unknown joint distribution $D$ on the product space $Z=X\times Y$. A learning algorithm is a mechanism that maps $S$ to a global map $f_S:X\rightarrow Y$ defined on the entire $X$. It is often assumed for simplicity that the algorithm is symmetric with respect to training sets---that the learning algorithm should return identical maps for two training sets with samples differing from each other only by permutation. We shall assume all maps considered here are measurable, and all measure spaces are separable. We are interested in the case where $X$ is a simple finite graph and $Y$ is the probability space $\P\left( N \right)$. The \emph{empirical risk} or \emph{empirical error} of a mapping $f_S:X\rightarrow Y$ learned from a training set $S$ of size $m>0$ is defined as
	\begin{equation*}
	R_m \left( f_S \right):=\frac{1}{m}\sum_{i=1}^mc \left( f_S, z_i \right)
	\end{equation*}
	where $c \left( \cdot,\cdot \right):Y^X\times \left( X\times Y \right)\rightarrow\mathbb{R}_{\geq0}$ is a cost function evaluating the predictive error of $f_S:X\rightarrow Y$ at a point sampled from the joint distribution $D$ on $X\times Y$. The \emph{generalization error} of the learned map is
	\begin{equation*}
	R_D \left( f_S \right)=\mathbb{E}_{z\sim D} \left[ c \left( f_S,z \right) \right]
	\end{equation*}
	which measures the average prediction error for a map learned from training data. The central problem in the PAC learning framework is bounding the discrepancy between $R_m$ and $R_D$. In \cite{Algorithmic_Stability}, the authors proved that such a bound exists if the algorithm satisfies a \emph{uniform stability} property, essentially meaning that the learned mapping changes very little in terms of predictive power if the training sample undergoes a small change.
	\begin{definition}[Uniform Stability, \cite{Algorithmic_Stability}]
		\label{defn:uniform-stability}
		Fix a positive integer $m\in\mathbb{Z}_+$. Let $S=\left\{ z_1,\cdots,z_m \right\}\subset X\times Y$ be a training set, and $S'$ be another training set that contains the same elements as $S$ with the only exception that the sample $z_i$ is replaced with a different sample $z_i'\neq z_i$. A learning algorithm $A:\left( X\times Y \right)^m\rightarrow Y^X$ that sends any training set $S$ to a mapping $f_S:X\times Y$ is said to be (uniform) $\beta$-stable for some positive constant $\beta>0$ if for any pair of training sets $S$, $S'$ that differ by exactly one element the following inequality holds:
		\begin{equation*}
		\left| c \left( f_S,z \right)-c\left(f_{S'},z\right) \right|\leq \beta\qquad\forall z\in X\times Y.
		\end{equation*}
	\end{definition}
	\begin{theorem}[\cite{Algorithmic_Stability}]
		\label{thm:bousquet-elisseeff}
		Let $S\mapsto f_S$ be a $\beta$-stable learning algorithm, such that $0\leq c \left( f_S,z \right)\leq M$ for all $z\in X\times Y$ and all learning set $S$. For any arbitrary $\epsilon>0$ we have for all $m\geq 8M^2/\epsilon^2$
		\begin{equation}
		\label{eq:fraction-bounds}
		\begin{aligned}
		\mathbb{P}_{S\sim D^m} &\left\{ \left| R_m \left( f_S \right)-R_D \left( f_S \right) \right| > \epsilon\right\}\leq \frac{64 Mm\beta+8M^2}{m\epsilon^2},
		\end{aligned}
		\end{equation}
		and for any $m\geq 1$
		\begin{equation}
		\label{eq:exponential-bounds}
		\begin{aligned}
		\mathbb{P}_{S\sim D^m}&\left\{ \left| R_m \left( f_S \right)-R_D \left( f_S \right) \right| > \epsilon+\beta\right\}\\
		&\qquad\qquad\leq 2\exp \left( -\frac{m\epsilon^2}{2 \left( m\beta+M \right)^2} \right).
		\end{aligned}
		\end{equation}
	\end{theorem}
	Of course, the order of $\beta$ in terms of training samples $m$ will be crucial here, otherwise any learning algorithm is uniformly stable for any bounded cost function. In \cite{Algorithmic_Stability} it was pointed out that a sufficient condition for these bounds to be tight is $\beta=O \left( 1/m \right)$ as $m\rightarrow\infty$. It was verified in \cite{Algorithmic_Stability} that the Tikhonov regularization framework for scalar-valued functions with quadratic cost function satisfies this requirement; but Theorem~\ref{thm:bousquet-elisseeff} is indeed much more general and applicable to any measurable spaces $X$ and $Y$. The rest of this paper is devoted to establishing algorithmic stability for(hyper)graph soft label propagation. 
	
	\subsection{Generalization bounds for Soft Label Propagation} 
	The goal of this subsection is to verify that the conditions of Theorem~\ref{thm:bousquet-elisseeff} are satisfied for the Tikhonov regularization framework \eqref{eq:tikhonov}.
	The first task is to find an appropriate model class for the distributions in $\P\left( N \right)$ that ensures uniform boundedness of the cost function
	\begin{equation}
	\label{eq:wasserstein-cost-func}
	c\left( f, \left( j,\mu_j \right) \right)=W_2^2 \left( f_j,\mu_j \right).
	\end{equation}
	This can be fulfilled trivially, for instance, if the metric space $\left( N,d_N \right)$ is of bounded diameter.
	This includes many generic applications we come across in practice, in particular for propagating histograms but are not already satisfied with popular distribution classes such as the Gaussian distribution. It is therefore preferable to work with a model class for distributions with uniformly bounded pairwise Wasserstein distances under mild assumptions. By definition \eqref{eq:wasserstein-distance}, bounding the Wassertein distance from above can be achieved by plugging an arbitrary coupling into the variational energy functional defining \eqref{eq:wasserstein-distance}. However, explicitly constructing meaningful couplings is typically difficult. Many existing bounds explore the multiscale structure of supports from the two distributions \cite{David1988,Lei2018,SP2018}, but it is not clear how those technical conditions can be used as model class specifications. Here we bypass this difficulty by leveraging the simple characterization of Wasserstein distances between one-dimensional distributions using quantile functions.
	
	According to Theorem~\ref{Thm:Wasserstein_One}, one can simplify \eqref{eq:tikhonov} as 
	\begin{equation*}
	\begin{aligned}
	\min_{f:V\to \P(N)}&\int_0^1 \Big[ \frac{1}{m} \sum_{i=1}^m \left(F_{\mu_i}^{-1}\left( s \right)-F_{f_i}^{-1}\left( s \right)\right)^2\\
	&\qquad +\gamma \sum_{\left( i,j \right)\in E}\left( F_{f_i}^{-1}\left( s \right)-F_{f_j}^{-1}\left( s \right) \right)^2  \Big]\mathrm{d}s.
	\end{aligned}
	\end{equation*}
	Because the inverse c.d.f.s and the distributions are in one-to-one correspondences, and all $F_{\mu_i}^{-1}$ are given, it suffices to solve for the $F_{f_i}^{-1}$'s in their entirety and then recover each probability distribution at vertex $i$ from $F_{f_i}^{-1}:[0,1]\rightarrow\mathbb{R}$. To simplify notation, we define $\Phi:V\times \left[ 0,1 \right]\rightarrow \mathbb{R}$ as $\Phi\left( i,s \right):=F_{f_i}^{-1}\left( s \right)$
	and denote $\Phi_s \left( i \right):=\Phi \left( i,s \right)$ for all $s\in \left[ 0,1 \right]$ and $i\in V$.
	For each fixed $s\in\left[0,1\right]$, $\Phi_s$ can be viewed as a function defined on vertices from graph $G$. For simplicity, we identify each $\Phi_s$ with a real column vector of length $n=\left| V \right|$. 
	Then the regularization term in \eqref{eq:tikhonov} can be written in terms of $L$, the weighted graph Laplacian of $G$. Thus \eqref{eq:tikhonov} transforms into
	\begin{equation}\label{eq:tikhonov-one-dimensional}
	\begin{aligned}
	&\min_{\Phi:V\times \left[ 0,1 \right]\rightarrow \mathbb{R}}\frac{1}{m}\sum_{i=1}^m\int_0^1 \left| F_{\mu_i}^{-1}\left( s \right)-\Phi_s \left( i \right) \right|^2\mathrm{d}s\\
	&\qquad \qquad\qquad \qquad +\gamma\int_0^1\Phi_s^{\top}L\Phi_s\,\mathrm{d}s.
	\end{aligned}
	\end{equation}
	The optimization problem \eqref{eq:tikhonov-one-dimensional} can be viewed as a linear combination of infinitely many Tikhonov regularization problems, one for each $s\in \left[ 0,1 \right]$ where each sub-problem is decoupled from others. Indeed, standard variational analysis shows that it suffices to solve each subproblem individually, i.e., solve for each fixed $s\in  \left[ 0,1 \right]$
	\begin{equation}
	\label{eq:tikhonov-one-dimensional-subproblem}
	\min_{\Phi_s\in\mathbb{R}^n}\frac{1}{m}\sum_{i=1}^m \left( F_{\mu_i}^{-1}\left( s \right)-\Phi_s \left( i \right) \right)^2+\gamma \Phi_s^{\top}L\Phi_s.
	\end{equation}
	Once all subproblems are solved, it is necessary to check compatibility across solutions $\left\{\Phi_s:s\in \left[ 0,1 \right]\right\}$, i.e., for any fixed $i\in V$, the map $s\mapsto \Phi_s \left( i \right)$ is indeed the inverse c.d.f. of a probability distribution. This compatibility will become straightforward after we derive the closed-form solution for each subproblem \eqref{eq:tikhonov-one-dimensional-subproblem}; see Proposition~\ref{Propo:Non_Decreasing} below.
	
	The solutions for Tikhonov regularization problems \eqref{eq:tikhonov-one-dimensional-subproblem} were known back in \cite{Belkin2004}. Let $\mathbf{1}=\left( 1,\cdots,1 \right)^{\top}\in\mathbb{R}^n$ be a column vector of all ones, and
	$$T_\ell=\mathrm{diag}\left( t_1,\cdots,t_{\ell},0,\cdots,0 \right)^{\top}\in\mathbb{R}^n$$
	where $t_i$ is the multiplicity of vertex $i\in V$ in the training set $S$ (we assumed without loss of generality that the training samples are the first $\ell$ vertices, for notational convenience), and
	\begin{equation}
	\label{eq:y-defn}
	\mathbf{y}_s=\left( \sum_{v_{i}=1}F_{\mu_i}^{-1}\left( s \right),\cdots,\sum_{v_i=\ell}F_{\mu_i}^{-1}\left( s \right),0,\cdots,0 \right)^{\top}\in\mathbb{R}^n
	\end{equation}
	i.e., for $1\leq i\leq \ell$, the $i$-th entry of $\mathbf{y}_s$ is the sum of the $t_i$ values of the inverse c.d.f.'s of $i\in V$. With this notation, it becomes easy to write down the Euler-Lagrange equation of the optimization problem \eqref{eq:tikhonov-one-dimensional-subproblem} as
	\begin{equation}
	\label{eq:tikhonov-one-dimensional-subproblem-euler-lagrange}
	\left( T_{\ell}+m\gamma L \right)\Phi_s^{*}=\mathbf{y}_s.
	\end{equation}
	To solve this equation, note that the operator  $T_{\ell}+m\gamma L$ may not be invertible---in fact, neither $T_{\ell}$ nor $L$ is invertible. Nevertheless, assuming the graph is connected, the nullspace of $L$ is one-dimensional and spanned precisely by the all-one vector $\mathbf{1}$. This means that $L$ will be invertible on the orthogonal complement of the one-dimensional subspace spanned by $\mathbf{1}$. Furthermore, noting that
	\begin{equation}
	\label{eq:standard-functional-analysis}
	T_{\ell}+m\gamma L=m\gamma \left( \frac{1}{m\gamma}T_{\ell}+L \right),
	\end{equation}
	by standard functional analysis (or \cite[Proof of Theorem 5]{Belkin2004}) we know that the perturbed operator $L+\left( m\gamma \right)^{-1}T_{\ell}$ is invertible on the orthogonal complement as well provided that $m\gamma$ is sufficiently large. More precisely, invertibility holds for
	\begin{equation*}
	\gamma\geq \frac{\max \left\{ t_1,\cdots,t_{\ell} \right\}}{m\lambda_1}
	\end{equation*}
	where $\lambda_1$ is the smallest non-zero eigenvalue of $L$, or the \emph{spectral gap} of the (possibly weighted) connected graph $G$. 
	This observation, together with the invariance of the quadratic cost in \eqref{eq:tikhonov-one-dimensional-subproblem} under global translations, allow us to preprocess the input data by subtracting scalar
	\begin{equation}
	\label{eq:off-set-defn}
	\bar{y}_s := \frac{1}{m}\mathbf{1}^{\top}\mathbf{y}_s=\frac{1}{m}\sum_{i=1}^mF_{\mu_i}^{-1}\left( s \right)
	\end{equation}
	from each $F_{\mu_i}^{-1}\left( s \right)$, applying the inverse of $T_\ell+m\gamma L$, and finally adding $\bar{y}_s$ back to the obtained solution. More specifically, we would like to solve the equivalent optimization problem
	\begin{equation}
	\label{eq:tikhonov-one-dimensional-subproblem-equiv}
	\begin{aligned}
	\Phi_s^*=\argmin_{\Phi_s\in\mathbb{R}^n}\frac{1}{m}&\sum_{i=1}^m \left[\left( F_{\mu_i}^{-1}\left( s \right)-\bar{y}_s\right)-\left(\Phi_s \left( i \right)-\bar{y}_s \right)\right]^2\\
	&+\gamma \left(\Phi_s-\bar{y}_s\mathbf{1}\right)^{\top}L\left(\Phi_s-\bar{y}_s\mathbf{1}\right),
	\end{aligned}
	\end{equation}
	which gives $\Phi_s^*-\bar{y}_s\mathbf{1}=\left( T_{\ell}+m\gamma L \right)^{-1}\left(\mathbf{y}_s-\bar{y}_sT_\ell\mathbf{1}\right).$
	Therefore, the solution to \eqref{eq:tikhonov-one-dimensional-subproblem} takes the form
	\begin{equation}
	\label{eq:sol-tikhonov-one-dimensional-subproblem}
	\begin{aligned}
	\Phi_s^{*}&=\left( T_{\ell}+m\gamma L \right)^{-1}\left(\mathbf{y}_s-\bar{y}_sT_\ell\mathbf{1}\right)+\bar{y}_s\mathbf{1}.
	\end{aligned}
	\end{equation}
	We emphasize here that the notation $\left( T_{\ell}+m\gamma L \right)^{-1}$ alone does not make sense because the matrix $T_{\ell}+m\gamma L$ may well be non-invertible; only the notation $\left( T_{\ell}+m\gamma L \right)^{-1}u$ for $u\in\mathbb{R}^n$ satisfying $\mathbf{1}^{\top}u=0$ bears meaning.
	\begin{remark}
		Alternatively, one can derive a solution to \eqref{eq:tikhonov-one-dimensional-subproblem} by directly applying the pseudo-inverse of $T_{\ell}+m\gamma L$ to $\mathbf{y}_s$, i.e., setting $\Phi_s^{*}:=\left( T_{\ell}+m\gamma L \right)^{\dagger}\mathbf{y}_s$. This avoids the requirement that $\gamma$ need not be too small, but leaves the algorithmic stability of the resulting solution $\Phi_s^{*}$ in question.
	\end{remark}
	
	Now that we have obtained closed-form solutions \eqref{eq:sol-tikhonov-one-dimensional-subproblem} to subproblems \eqref{eq:tikhonov-one-dimensional-subproblem} for each $s\in \left[ 0,1 \right]$, it is imperative to guarantee that the closed-form solutions $\left\{ \Phi_s^{*}\mid 0\leq s\leq 1 \right\}$ piece together and give rise to inverse c.d.f.'s at each vertex $i\in V$. This requires that, for each $i\in V$, the map $\left[ 0,1 \right]\ni s\mapsto \Phi_s^{*}\left( i \right)\in\mathbb{R}$ should be non-decreasing and right continuous. The right continuity is obvious, because for each $i\in V$ the map $\left[ 0,1 \right]\ni s\mapsto \mathbf{y}_s \left( i \right)$ is right continuous, and the linear combination of right continuous functions is still right continuous, thus this assertion follows from the closed-form expression \eqref{eq:sol-tikhonov-one-dimensional-subproblem}. Monotonicity would be guaranteed if there is a ``maximum principle'' for the operator $T_{\ell}+m\gamma L$, or equivalently $L+\left( m\gamma \right)^{-1}T$, on the graph $G$, i.e., if $\mathbb{R}^{n}\ni \mathbf{y}\geq 0$ (entrywise) and $\left( T_{\ell}+m\gamma L \right)\Phi=\mathbf{y}$ then $\Phi\geq 0$ (entrywise). This is because we already have $\mathbf{y}_s-\mathbf{y}_t\geq 0$ for any $0\leq t\leq s\leq 1$ by the monotonicity of the inverse c.d.f.'s, hence such a ``maximum principle'' would guarantee $\Phi_s-\Phi_t\geq 0$ (entrywise). Such maximum principles abound for graph Laplacians, see e.g., \cite{HS1997,CCK2007}. It is natural to expect such a maximum principle to hold for $L+\left( m\gamma \right)^{-1}T$ as well, since $T$ is a non-negative.
	
	\begin{lemma}[Maximum Principle]
		\label{lem:maximum-principle}
		If $\Phi\in\mathbb{R}^n$ is such that $\left[\left( T_{\ell}+m\gamma L \right)\Phi\right] \left(i\right)\geq 0$ for all $1\leq i\leq \ell$ and $\left[\left( T_{\ell}+m\gamma L \right)\Phi\right] \left(i\right)=0$ for all $\ell+1\leq i\leq n$, then $\Phi$ attains both its maximum and minimum over $i=1,\cdots,n$ within $\left\{ 1,\cdots,\ell \right\}$. In particular, $\Phi\left(i\right)\geq 0$ for all $1\leq i\leq n$.
	\end{lemma}
	\begin{proof}
		The conditions on $\Phi$ can be written as
		\begin{align}
		\left[\frac{t_i}{m\gamma}+\mathrm{deg}\left(i\right)\right]\Phi \left( i \right)-\sum_{j:j\sim i}\Phi \left( j \right)\geq 0 & \qquad 1\leq i\leq \ell\label{eq:maximum-principle-boundary}\\
		\mathrm{deg}\left(i\right)\Phi \left( i \right)-\sum_{j:j\sim i}\Phi \left( j \right)=0 & \qquad \ell+1\leq i\leq n\label{eq:maximum-principle-interior}
		\end{align}
		where $\mathrm{deg}\left( i \right)\geq 1$ is the degree of vertex $i$ in graph $G$. First, we assert that the minimum of $\Phi$ must be attained among the vertices $1,\cdots,\ell$, for otherwise, if $\ell+1\leq i_*=\argmin_{i\in V}\Phi \left( i \right)\leq n$, then by \eqref{eq:maximum-principle-interior} we have
		
		\begin{equation*}
		\begin{aligned}
		\mathrm{deg}\left(i_*\right)\Phi \left( i_* \right)&=\sum_{j:j\sim i_*}\Phi \left( j \right)\\
		&\geq \sum_{j:j\sim i_*}\Phi \left( i_* \right)=\mathrm{deg}\left(i_*\right)\Phi \left( i_* \right)
		\end{aligned}
		\end{equation*}
		which implies $\Phi \left( j \right)=\Phi \left( i_* \right)$ for all vertices $j\sim i_*$. This argument can be repeated until the constant value propagates into the vertices within $1,\cdots,\ell$, and the assertion follows from the connectivity of the graph. The assertion for the maximum can be established analogously. Next we argue that the minimum of $\Phi$ on the vertices of $G$ must be non-negative. Assume the contracy, i.e. the minimum attained at $i_*\in \left[ 1,\ell \right]$ is strictly negative, then by \eqref{eq:maximum-principle-boundary} we have
		\begin{equation*}
		\begin{aligned}
		0 &\leq \left[\frac{t_{i_*}}{m\gamma}+\mathrm{deg}\left(i_{*}\right)\right]\Phi \left( i_{*} \right)-\sum_{j:j\sim i_{*}}\Phi \left( j \right)\\
		&=\frac{t_{i_*}}{m\gamma}\Phi \left( i_* \right)+\sum_{j:j\sim i_*}\left[ \Phi \left( i_* \right)-\Phi \left( j \right) \right]<0
		\end{aligned}
		\end{equation*}
		where the strict inequalty follows from the counter-assumption $\Phi \left( i_{*} \right)<0$. This contradiction completes our proof that $\Phi \geq 0$ on the entire graph $G$.
		
	\end{proof}
	This lemma then implies the promised monotonicity.  
	\begin{proposition}\label{Propo:Non_Decreasing}
		For any vertex $i\in V$, the closed-form solutions \eqref{eq:sol-tikhonov-one-dimensional-subproblem} is non-decreasing with respect to $s\in \left[ 0,1 \right]$.
	\end{proposition}
	\begin{proof}
		By the equivalence of \eqref{eq:tikhonov-one-dimensional-subproblem-equiv} and \eqref{eq:tikhonov-one-dimensional-subproblem}, solutions $\Phi_s$ satisfy the Euler-Lagrange equations for \eqref{eq:tikhonov-one-dimensional-subproblem}:
		\begin{equation*}
		\left( T_{\ell}+m\gamma L \right)\Phi_s^{*}=\mathbf{y}_s.
		\end{equation*}
		For any $0\leq t\leq s\leq 1$, subtracting two Euler-Lagrange equations yields
		\begin{equation*}
		\left( T_{\ell}+m\gamma L \right)\left(\Phi_s^{*}-\Phi_t^{*}\right)=\mathbf{y}_s-\mathbf{y}_t\geq 0
		\end{equation*}
		where the inequality follows from the definition of $\mathbf{y}_s$ in \eqref{eq:y-defn}. Furthermore, it is straightforward to see that $\mathbf{y}_s-\mathbf{y}_t$ satisfies the assumption in Lemma~\ref{lem:maximum-principle}, which then implies $\Phi_s^{*}\geq \Phi_t^{*}$.
	\end{proof}
	
	We can now rest assured that the solutions \eqref{eq:sol-tikhonov-one-dimensional-subproblem} constitute an inverse c.d.f. at each vertex $i\in V$. But there is more: it can be easily verified that \eqref{eq:tikhonov-one-dimensional-subproblem-equiv} is equivalent to the Tikhonov regularization problem formulated in \cite{Belkin2004} if we view $\left(\Phi_s-\bar{y}_s\mathbf{1}\right)$ as variables. We can thus follow the idea of \cite[Theorem 5]{Belkin2004} to get algorithmic stability for each individual $\Phi_s$, $s\in \left[ 0,1 \right]$.
	
	\begin{theorem}
		\label{thm:slice-algorithmic-stability}
		Assume $m\geq 4$ and $0<T:=\max \left\{ t_1,\cdots,t_{\ell} \right\}<\infty$ satisfies $m\gamma\lambda_1-T>0$, where $\lambda$ is the regularization parameter in \eqref{eq:tikhonov-one-dimensional-subproblem} and $\lambda_1$ is the spectral gap of the connected graph $G$. Let $S=\left\{ \left( v_i,\mu_i \right)\mid 1\leq i\leq m,\,\,v_i\in V,\,\,\mu_i\in\P\left( \mathbb{R} \right) \right\}$ and $S'=\left\{ \left( v'_i,\mu'_i \right)\mid 1\leq i\leq m,\,\,v_i\in V,\,\,\mu_i\in\P\left( \mathbb{R} \right) \right\}$ be two training sets that differ from each other by exactly one data sample. Assume further that, for a fixed $s\in \left[ 0,1 \right]$ there holds
		\begin{equation}
		\label{eq:quantile-boundedness}
		\max \left\{ \left|F_{\mu_i}^{-1} \left( s \right)\right|, \left|F_{\mu_i'}^{-1} \left( s \right)\right|,\,\,i=1,\cdots,m\right\}\leq M_s<\infty.
		\end{equation}
		Let $\Phi_s^{*},\Phi_s'^{*}$ be solutions of \eqref{eq:tikhonov-one-dimensional-subproblem} for $S$ and $S'$, respectively,
		\begin{equation*}
		\begin{aligned}
		\Phi_s^{*}&=\left( T_{\ell}+m\gamma L \right)^{-1}\left(\mathbf{y}_s-\bar{y}_sT_\ell\mathbf{1}\right)+\bar{y}_s\mathbf{1}\\
		\Phi_s'^{*}&=\left( T_{\ell}'+m\gamma L \right)^{-1}\left(\mathbf{y}'_s-\bar{y}'_sT'_\ell\mathbf{1}\right)+\bar{y}_s'\mathbf{1}
		\end{aligned}
		\end{equation*}
		where $T_{\ell}'$, $\mathbf{y}'_s$, $\bar{y}_s'$ are defined analogously to $T_{\ell}$, $\mathbf{y}_s$, $\bar{y}_s$ but with respect to $S'$ instead of $S$. Then
		\begin{equation}
		\label{eq:slice-algorithmic-stability}
		\left\| \Phi_s^{*} -\Phi_s'^{*} \right\|_{\infty} \leq \frac{3M_s\sqrt{Tm}}{\left( m\gamma\lambda_1-T \right)^2}+\frac{4M_s}{m\gamma\lambda_1-T}+\frac{2M_s}{m}.
		\end{equation}
	\end{theorem}
	\begin{proof}
		Following the same argument as in the proof of \cite[Theorem 5]{Belkin2004}, we can assume without loss of generality that $S$, $S'$ differ by a new point $\left( v_m,\mu_m \right)\leftrightarrow \left( v_m',\mu_m' \right)$; the other case where only the multiplicities differ can be treated similarly. By our assumption \eqref{eq:quantile-boundedness}, the two averages differ by at most an amount of
		\begin{equation*}
		\left| \bar{y}_s-\bar{y}_s' \right|\leq \frac{2M_s}{m}.
		\end{equation*}
		For simplicity, introduce temporary notations
		\begin{equation*}
		A:=T_{\ell}+m\gamma L,\qquad B:=T_{\ell}'+m\gamma L.
		\end{equation*}
		Using the simple fact that the $2$-norm dominate the $\infty$-norm, we have
		\begin{equation*}
		\begin{aligned}
		&\left\| \Phi_s^{*} -\Phi_s'^{*} \right\|_{\infty} \leq \left\| \Phi_s^{*} -\Phi_s'^{*} \right\|_2\\
		&\leq \frac{2M_s}{m}+\left\| A^{-1}\left(\mathbf{y}_s-\bar{y}_sT_\ell\mathbf{1}\right)-B^{-1}\left(\mathbf{y'}_s-\bar{y}_s'T'_\ell\mathbf{1}\right) \right\|_2\\
		&\leq \frac{2M_s}{m}+\left\| A^{-1}\left(\mathbf{y}_s-\bar{y}_sT_\ell\mathbf{1}\right)-A^{-1}\left(\mathbf{y'}_s-\bar{y}_s'T'_\ell\mathbf{1}\right) \right\|_2\\
		&\qquad+\left\| A^{-1}\left(\mathbf{y'}_s-\bar{y}_s'T'_\ell\mathbf{1}\right)-B^{-1}\left(\mathbf{y'}_s-\bar{y}_s'T'_\ell\mathbf{1}\right) \right\|_2.
		\end{aligned}
		\end{equation*}
		Standard functional analysis argument (the same perturbation reasoning we gave in \eqref{eq:standard-functional-analysis}) tells us that $\left\| A^{-1} \right\|_2\leq \left( m\gamma\lambda_1-T \right)^{-1}$. Together with the observation that
		\begin{align*}
		&\left\| \left(\mathbf{y}_s-\bar{y}_sT_\ell\mathbf{1}\right) - \left(\mathbf{y'}_s-\bar{y}_s'T'_\ell\mathbf{1}\right) \right\|_2\\
		&\leq \left\| \mathbf{y}_s-\mathbf{y}_s' \right\|_2+\left\| \bar{y}_sT_\ell\mathbf{1}-\bar{y}_s'T'_\ell\mathbf{1} \right\|_2\\
		&\leq 2M_s+\frac{2M_s}{m}<4M_s
		\end{align*}
		we have
		\begin{equation*}
		\left\| A^{-1}\left(\mathbf{y}_s-\bar{y}_sT_\ell\mathbf{1}\right)-A^{-1}\left(\mathbf{y'}_s-\bar{y}_s'T'_\ell\mathbf{1}\right) \right\|_2\leq \frac{4M_s}{m\gamma\lambda_1-T}.
		\end{equation*}
		In the meanwhile, noting that we also have $\left\| B^{-1} \right\|_2\leq \left( m\gamma\lambda_1-T \right)^{-1}$, and $\left\| A-B \right\|_2=\left\| T_{\ell}'-T_{\ell} \right\|_2 \leq \sqrt{2}<3/2$, we conclude that
		\begin{equation*}
		\begin{aligned}
		&\left\| A^{-1}\left(\mathbf{y'}_s-\bar{y}_s'T'_\ell\mathbf{1}\right)-B^{-1}\left(\mathbf{y'}_s-\bar{y}_s'T'_\ell\mathbf{1}\right) \right\|_2\\
		&=\left\| B^{-1} \left( B-A \right) A^{-1}\left(\mathbf{y'}_s-\bar{y}_s'T'_\ell\mathbf{1}\right)\right\|_2\leq \frac{3M_s\sqrt{Tm}}{\left( m\gamma\lambda_1-T \right)^2}.
		\end{aligned}
		\end{equation*}
		Putting everything together completes the proof.
		
	\end{proof}

	The boundedness assumption on $\Phi_s$ seems artificial, but is actually natural: an almost identical argument as the first part of the proof of Lemma~\ref{lem:maximum-principle}, with minimum replaced with maximum and \emph{mutatis mutandis}, establishes that the global maximum of $\Phi_s$ must be attained at the boundary $1\leq i\leq \ell$.  Hence, because there are only finitely many data in the training set, this boundedness is a mild requirement (e.g., satisfied if each $F_{\mu_i}^{-1} \left( s \right)$ is finite). We define a model class to reflect the requirement that the inverse c.d.f.'s of one-dimensional probability distributions in the training set should be controlled. We define the model class in Definition~\ref{defn:dqc} and summarize the maximum principle argument as a lemma on \emph{a priori} estimates for future convenience. 
	
	\begin{definition}[Dominated Quantile Class]
		\label{defn:dqc}
		Let $\phi\in L^2 \left[ 0,1 \right]$ and $\phi\geq 0$ on $\left[ 0,1 \right]$. A probability distribution $\mu\in\P\left( \mathbb{R} \right)$ is said to belong to \emph{dominated quantile class $\mathcal{M}_{\phi}^2$} if $\left|F_{\mu}^{-1}\left( s \right)\right|\leq \phi \left( s \right)$ for e.g., $s\in \left[0,1\right]$.
	\end{definition}
	
	\begin{lemma}[\emph{A Priori} Estimates]
		\label{lem:apriori-estimates}
		If in the training set $S=\left\{ \left( v_i,\mu_i \right)\mid 1\leq i\leq m,\,\,v_i\in V,\,\,\mu_i\in\P\left( \mathbb{R} \right) \right\}$ all $\mu_i$ lie in a dominated quantile model class $\mathcal{M}_{\phi}^2$ for some $\phi\in L^2 \left[ 0,1 \right]$ with $\phi\geq 0$ on $\left[ 0,1 \right]$, then any map $f:V\rightarrow\P\left( \mathbb{R} \right)$ minimizing \eqref{eq:tikhonov} takes values in $\mathcal{M}_{\phi}^2$ as well.
	\end{lemma}
	\begin{proof}
		By the equivalence between \eqref{eq:tikhonov} and \eqref{eq:tikhonov-one-dimensional}, it suffices to show the following fact: for each fixed $s\in \left[ 0,1 \right]$, if $\max \left\{ \left|F_{\mu_i}^{-1} \left( s \right)\right|,\,\,i=1,\cdots,m\right\}\leq \phi \left( s \right)$ then $\left\|\Phi_s^{*}\right\|_{\infty}\leq \phi \left( s \right)$, where $\Phi_s^{*}$ is defined in \eqref{eq:tikhonov-one-dimensional-subproblem-equiv}. But this follows straightforwardly from the maximum principle.
	\end{proof}

	We now present the main theoretical result of this paper. In our setting these results apply to graphs as well as hypergraphs by Proposition~\ref{prop:clique-equivalence}.
	
	\begin{proposition}[Algorithmic Stability for Soft Label Propagation of One-Dimensional Distributions]
		\label{prop:alg-stability-slp}
		Assume $m\geq 4$ and $0<T:=\max \left\{ t_1,\cdots,t_{\ell} \right\}<\infty$ satisfying $m\gamma\lambda_1-T>0$, where $\gamma$ is the regularization parameter in \eqref{eq:tikhonov-one-dimensional-subproblem} and $\lambda_1$ is the spectral gap of the weighted, connected graph $G$. If the joint distribution $D\in\P\left( V\times \P\left( \mathbb{R} \right) \right)$ is supported on $V\times \mathcal{M}_{\phi}^2$ for a quantile model class $\mathcal{M}_{\phi}^2\subset \P\left( \mathbb{R} \right)$ for some $\phi\in L^2 \left[ 0,1 \right]$ with $\phi\geq 0$ on $\left[ 0,1 \right]$, then the solutions of \eqref{eq:tikhonov} or \eqref{eq:tikhonoc_Hyper} are $\beta$-stable in the sense of Definition~\ref{defn:uniform-stability} with respect to cost function \eqref{eq:wasserstein-cost-func}, where
		\begin{equation}
		\label{eq:beta-expression}
		\beta=4\left\| \phi \right\|_2^2\left[\frac{3\sqrt{Tm}}{\left( m\gamma\lambda_1-T \right)^2}+\frac{4}{m\gamma\lambda_1-T}+\frac{2}{m}\right].
		\end{equation}
	\end{proposition}
	
	\begin{proof}
		Let $\left( j,\theta_j \right)$ be a new sample drawn from the joint distribution $D$. Then $\theta_j\in\mathcal{M}_{\phi}^2$ with probability $1$. Let $S$, $S'$ be two training samples with values in $\mathcal{M}_{\phi}^2$ and differ by exactly one data point. By Theorem~\ref{thm:slice-algorithmic-stability} we have
		\begin{equation}
		\label{eq:sliced-bounds}
		\begin{aligned}
		&\left| \Phi_s^{*} \left( j \right) -\Phi_s'^{*} \left( j \right) \right| \\
		&\leq \left[\frac{3\sqrt{Tm}}{\left( m\gamma\lambda_1-T \right)^2}+\frac{4}{m\gamma\lambda_1-T}+\frac{2}{m}\right]\phi \left( s \right).
		\end{aligned}
		\end{equation}
		By \eqref{Eq:One_D_Equivalence}, the difference between the squared Wasserstein losses satisfy
		\begin{equation*}
		\begin{aligned}
		&\left| c \left( f_S, \left( j,\theta_j \right) \right) - c \left( f_{S'}, \left( j,\theta_j \right) \right) \right|\\
		&=\left| W_2^2 \left( f_S \left( j \right),\theta_j \right) - W_2^2 \left( f_{S'} \left( j \right),\theta_j \right) \right|\\
		&=\left| \int_0^1 \left| \Phi_s^{*} \left( j \right)-F_{\theta_j}^{-1}\left( s \right) \right|^2\mathrm{d}s - \int_0^1 \left| \Phi_s'^{*} \left( j \right)-F_{\theta_j}^{-1}\left( s \right) \right|^2\mathrm{d}s \right|\\
		&\leq \int_0^1 \left| \left( \Phi_s^{*} \left( j \right)+\Phi_s'^{*} \left( j \right)-2F_{\theta_j}^{-1}\left( s \right) \right)\left( \Phi_s^{*} \left( j \right) -\Phi_s'^{*} \left( j \right) \right) \right| \mathrm{d}s\\
		&\stackrel{\left( * \right)}{\leq} \left[\frac{3\sqrt{Tm}}{\left( m\gamma\lambda_1-T \right)^2}+\frac{4}{m\gamma\lambda_1-T}+\frac{2}{m}\right]\cdot \int_0^1 4\phi \left( s \right)\cdot \phi \left( s \right)\mathrm{d}s\\
		&=4\left\| \phi \right\|_2^2\left[\frac{3\sqrt{Tm}}{\left( m\gamma\lambda_1-T \right)^2}+\frac{4}{m\gamma\lambda_1-T}+\frac{2}{m}\right]=\beta,
		\end{aligned}
		\end{equation*}
		where at $\left( * \right)$ we used \eqref{eq:sliced-bounds} to bound the difference $\left| \Phi_s^{*} \left( j \right) -\Phi_s'^{*} \left( j \right) \right|$, and invoked Lemma~\ref{lem:apriori-estimates} to conclude that
		\begin{equation*}
		\Phi_s^{*}\left( j \right),\Phi_s'^{*}\left( j \right) \leq \phi \left( s \right)
		\end{equation*}
		and hence
		\begin{equation*}
		\left| \Phi_s^{*} \left( j \right)+\Phi_s'^{*} \left( j \right)-2F_{\theta_j}^{-1}\left( s \right) \right|\leq 4\phi \left( s \right). \qedhere
		\end{equation*}
	\end{proof}
	
	Note that the cost function is uniformly bounded by $M=4 \left\| \phi \right\|_2^2$ in our setting. Our main result follows from combining  Proposition~\ref{prop:alg-stability-slp} and Theorem~\ref{thm:bousquet-elisseeff}.
	\begin{theorem}[Generalization Error for Soft Label Propagation for One-Dimensional Distributions]\label{Thm:Gen_Error_Soft}
		Under the same assumptions as Proposition~\ref{prop:alg-stability-slp}, for any $\epsilon>0$ we have for all $m\geq 8M^2/\epsilon^2$
		\begin{equation}
		\label{eq:fraction-bounds-1}
		\mathbb{P}_{S\sim D^m} \left\{ \left| R_m \left( f_S \right)-R_D \left( f_S \right) \right| > \epsilon\right\}\leq \frac{64 Mm\beta+8M^2}{m\epsilon^2},
		\end{equation}
		and for any $m\geq 1$
		\begin{equation}
		\label{eq:exponential-bounds-1}
		\begin{aligned}
		\mathbb{P}_{S\sim D^m}&\left\{ \left| R_m \left( f_S \right)-R_D \left( f_S \right) \right| > \epsilon+\beta\right\}\\
		&\leq 2\exp \left( -\frac{m\epsilon^2}{2 \left( m\beta+M \right)^2} \right),
		\end{aligned}
		\end{equation}
		where $M=4\left\| \phi \right\|_2^2$ and $\beta$ given by \eqref{eq:beta-expression}.
	\end{theorem}
	
	\section{Numerical Experiments}
\subsection{Label Propagation Algorithm}
Alg.~\ref{alg:AlternatingLabelPropagation} details the label propagation algorithm we use to obtain the results in the next two sections. 
\def\hs{4mm}
\begin{algorithm}[ht]
	\SetAlgoLined
	\DontPrintSemicolon
	\LinesNumbered
	\KwData{hypergraph $H = (V, \mathcal{E})$; a subset of vertices $V_0$ with known labels $\bar{l}(v)$, $\forall v\in V_0$; parameters $\alpha, \gamma > 0$, a condition $\mathsf{EC}$ for exiting the main loop on line \ref{algline:mainLoop}.}
	\KwResult{labels $l(v)$, $\forall v\in V$.}
	Randomly initialize labels $l(v)$, $\forall v\in V$\;
	\For{every $E\in \mathcal{E}$}{
		\For{every $v\in E$}{
			\eIf{$v \in V_0$}{
				$W_E(v) = \alpha$\; \label{algline:alpha}
			}{
				$W_E(v) = 1$\;
			}
		}
	}
	\For{every $v\in V$}{
		\For{every $E\in\mathcal{E}$ incident to $v$}{
			$w_v(E) = 1/|E|$\;
		}
		\If{every $v \in V_0$}{
			append vector $W_v$ with $\gamma$\;
		}
	}
	\While{$\mathsf{EC}$ is not met\label{algline:mainLoop}}{
		Initialize $loss = 0$\;
		\For{every $E\in \mathcal{E}$\label{algline:bCenterforEdge_start}}{
			$L_E = (l(v))_{v\in E}$\;
			$l(E) = \mathsf{Barycenter}\paren{W_E, L_E}$\;
		}\label{algline:bCenterforEdge_end}
		\For{every $v\in V$\label{algline:bCenterforVertex_start}}{
			$L_v = \paren{l(E)}_{\textrm{$E$ incident to $v$}}$\;
			\If{$v\in V_0$}{
				append $L_v$ with $\bar{l}(v)$
			}
			$l(v) = \mathsf{Barycenter}\paren{W_v, L_v}$\;
			\For{every $E\in\mathcal{E}$ incident to $v$}{
				$loss = loss + W_v(E) \cdot \mathsf{WassDist}(l(v), l(E))$
			}
			\If{$v \in V_0$}{
				$loss = loss + \gamma \cdot \mathsf{WassDist}\paren{l(v), \bar{l}(v)}$
			}
		}\label{algline:bCenterforVertex_end}
	}
	\caption{Alternating label propagation algorithm}
	\label{alg:AlternatingLabelPropagation}
\end{algorithm}

The functions $\mathsf{Barycenter}$ and $\mathsf{WassDist}$ can be any algorithms that calculate the weighted Wasserstein barycenter of a vector of labels $L$ with weights $W$, and the Wasserstein distance between two input labels, respectively. Note that we introduce another parameter $\alpha>1$ to adjust the weights of vertices with known labels (in line \ref{algline:alpha}) in order to increase their influences to hyperedge barycenters. Similar techniques are explored in \cite{shi2017weighted,shi2018generalization}.

The algorithm relies on the alternating technique in minimizing \eqref{eq:tikhonoc_Hyper} in each iteration. This technique consists of two steps:  (i) first calculates the barycenters $\mathsf{bar}(E)$ of all hyperedges $E$ using the current labels of vertices they contain and treats the derived barycenters as the labels of the hyperedges (lines \ref{algline:bCenterforEdge_start} to \ref{algline:bCenterforEdge_end}), and (ii) then calculates the barycenters, i.e.~the new labels, of all vertices using labels of the hyperedges incident to them, together with their targeted labels if the latter are known (line \ref{algline:bCenterforVertex_start} to \ref{algline:bCenterforVertex_end}). Due to the alternating nature of the algorithm, we call it  
\textit{alternating label propagation}.

\subsection{Stochastic Block Model}
In the first two experiments, we run label propagation on $3$-uniform hypergraphs generated using the stochastic block model (SBM) over $100$ vertices that are grouped into either $2$ or $3$ blocks. More specifically, the probability that a hyperedge $\{v_i, v_j, v_r\}$ exists is $p = 0.01$ if all $v_i, v_j$, and $v_r$ belong to the same block and is  $q = 0.002$ otherwise. 

We set the soft labels to be $b$-dimensional Gaussian distributions, where $b$ is the number of blocks. For any vertex from block $i$, $i = 1,\dots, b$, whose label is known, we set the mean of its label to be $e_{i}$, where $e_{i}$ is the base vector with the $i$-th coordinate being $1$ and the rest being $0$. The covariance matrix of each known label is set to be $0.05I_b$, where $I_b$ is the $b$-dimensional identity matrix. The predicted block assignment of a vertex is the $\argmax$ of its predicted mean. In both of the experiments, we use $\alpha = 20$ and $\gamma = 10$. We run the experiments with $5$ to $15$ vertices of known block assignment from each block, and the error bars are obtained by averaging over $20$ random selections of vertices with known labels. 

\begin{figure}[ht]
	\centering
	\subfloat[SBM with $2$ blocks]{\includegraphics[scale=.33]{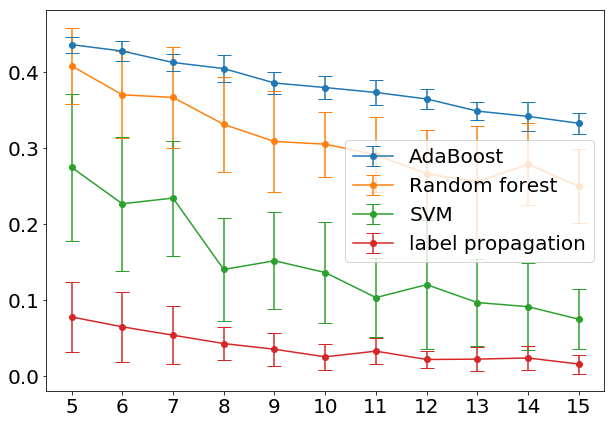}\label{subfig:SMB_2}} \qquad \qquad \qquad
	\subfloat[SBM with $3$ block]{\includegraphics[scale=.33]{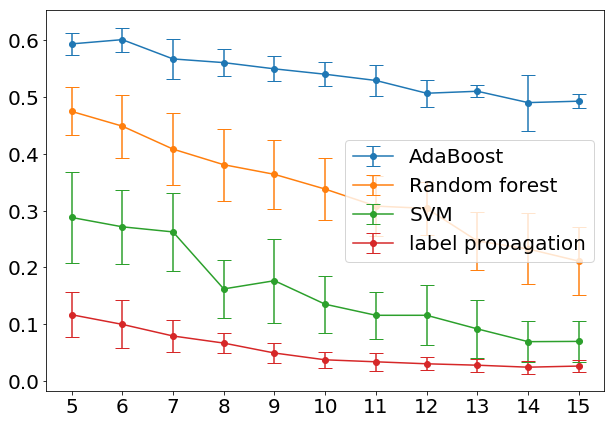}\label{subfig:SMB_3}}
	\caption{Comparison of traditional classification algorithms with hypergraph label propagation on SBM.}
	\label{fig:SBM}
\end{figure}

We compare the performance of our label propagation approach with with AdaBoost, random forest, and SVM  in Fig.~\ref{fig:SBM}. We use incidence matrix as the feature matrix in  AdaBoost, Random forest, and SVM to solve the classification problem. 

\noindent\textbf{SBM with two blocks:}
The hypergraph generated for this experiment has two blocks of sizes $50$ and $50$, and $629$ hyperedges with $388$ of them containing vertices from one block.

\noindent\textbf{SBM with three blocks:}
The hypergraph generated for this experiment has three blocks of sizes $33$, $33$, and $34$, and $384$ hyperedges with $182$ of them has vertices from one block. 

\subsection{UCI datasets}
In the next two experiments, we apply our label propagation as a classification algorithm to the following two datasets with categorical features from the UCI machine learning repository:

\noindent\textbf{Congressional Voting Records:} This dataset contains voting records on $16$ issues of the $2$nd session of the $98$th Congress. We form a pair of hyperedges for each issue each of which contains voters who voted "Yay" and "Nay", respectively. For voters whose votes were missing, we don't include them in any of the hyperedges constructed for the corresponding issue. This resilience to the missing data samples illustrates another advantage of applying hypergraph label propagation to classification problems. We test label propagation algorithm with $5$, $10$, $15$, $20$, $25$, and $30$ congressmen and women from each party whose affiliation are given.
\vspace{1mm}

\noindent\textbf{Mushrooms:} This dataset contains $22$ features (e.g., shapes, colors, and habitats, etc) of $8124$ mushrooms. We form $97$ hyperedges each of which contains mushrooms sharing identical features. We choose $1000$ edible and $1000$ poisonous mushrooms to run the experiment. We run the algorithm in 6 cases where $10$, $20$, $30$, $40$, $50$, and $60$ mushrooms are given labels from each category.

\begin{figure}[ht]
	\centering
	\subfloat[Congressional voting records]{\includegraphics[scale=.33]{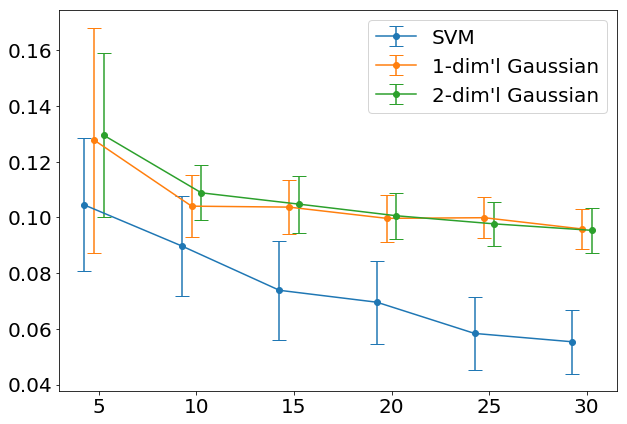}\label{subfig:Congress}}\qquad \qquad \qquad
	\subfloat[Mushrooms]{\includegraphics[scale=.33]{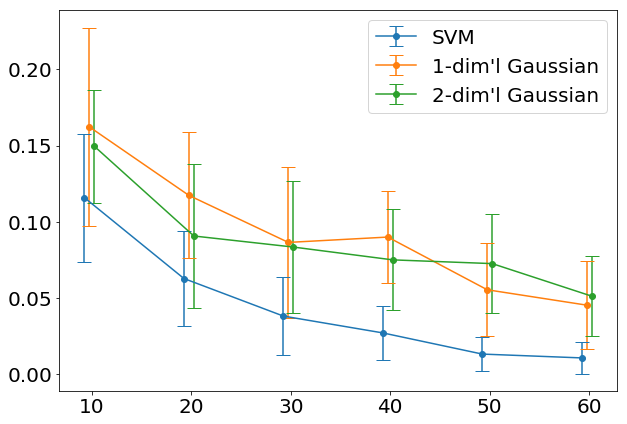}\label{subfig:Mushroom}}
	\caption{Comparison of SVM with hypergraph label propagation as a classification algorithm.}
	\label{fig:UCI_Congress_Mushroom}
\end{figure}

In both datasets, the soft labels are either $1$-dimensional Gaussian distributions $N(+1, 0.01)$ and $N(-1, 0.01)$ or $2$-dimensional Gaussian distributions $N((1, 0), 0.01I_2)$ and $N((0, 1), 0.01I_2)$ depending on which class the labelled sample belongs to. 
The predicted class of a vertex is obtained as follows: For the $1$-dimensional case, it is the sign of the mean of its label and for the $2$-dimensional case, it is $+1$ if the first coordinate of the mean vector of its label is larger than the second coordinate and $-1$ otherwise. 
For both experiments, we set $\alpha = 10$ and $\gamma = 1$. The error bars are obtained by averaging $20$ random selections of vertices with known labels. 
We compare the performance of hypergraph label propagation (as a classification algorithm) with SVM in Fig.~\ref{fig:UCI_Congress_Mushroom}. 

\subsection{Discussion of numerical experiments} 
The above experiments demonstrate that the hypergraph label propagation can serve as a powerful alternative classification algorithm especially when the dataset is structured as a network (for example as in SBM). 
The reason as to why the traditional classification algorithms may fail on network-like datasets (as illustrated in Fig.~\ref{fig:SBM}) is because for these datasets almost all coordinates of a feature vector tend to be identical except for few of them. We can understand these features as describing only local properties of the dataset. Therefore, they can give rise to global characterizations of the datasets, in a substantial way, only when properly ``patched'' together. Label propagation algorithm provides a novel way of combining features which is shown in Fig.~\ref{fig:SBM} to outperform the classical algorithms.

	\section{Conclusion}
	In this paper, we proposed a novel framework for a semi-supervised learning problem where (i) the labels are given by probability measures on a metric space (``soft labels'') and (ii) the underlying similarity structure is given by a hypergraph, which subsumes graphs and simplicial complexes. Our framework was inspired by a re-formulation of graph-based label propagation in terms of message passing and borrowed ideas from the theory of multi-marginal optimal transport. We then established generalization error bounds for propagating one-dimensional distributions using $2$-Wasserstein distances. To the best of our knowledge, this constitutes the first generalization error bounds for Wasserstein distance based soft label propagation, even on graphs. We expect similar generalization bounds to hold for propagating higher-dimensional probability distributions as well as using other Wasserstein distances, but a deeper understanding of the geometry underlying Wasserstein spaces will be indispensable for those purposes. Future work includes (i) generalization of our results to higher-dimensional probability measures, (ii) investigating the scalability and efficiency of our message-passing algorithm, and (iii) experimental study of our framework on real-work networks that can be naturally represented by hypergraphs.

	\bibliography{bibliography}
	\bibliographystyle{alpha}

\end{document}